\documentclass[sigconf]{acmart}

\setcopyright{none}
\acmYear{}
\copyrightyear{}
\acmConference{}
\acmBooktitle{}
\acmDOI{}
\acmISBN{}

\usepackage{tikz}
\usetikzlibrary{trees}
\usepackage[ruled,vlined]{algorithm2e}

\begin{document}

\title{Hierarchical Contextual Uplift Bandits for Catalog Personalization}
\subtitle{Preprint}

\author{Anupam Agrawal}
\email{anupam.agrawal@dream11.com}
\affiliation{%
  \institution{Dream11}
  \city{Mumbai}
  \state{Maharashtra}
  \country{India}
}

\author{Rajesh Mohanty}
\email{rajesh.mohanty@dream11.com}
\affiliation{%
  \institution{Dream11}
  \city{Mumbai}
  \state{Maharashtra}
  \country{India}
}

\author{Shamik Bhattacharjee}
\email{shamik.bhattacharjee@dream11.com}
\affiliation{%
  \institution{Dream11}
  \city{Mumbai}
  \state{Maharashtra}
  \country{India}
}

\author{Abhimanyu Mittal}
\email{abhimanyu.mittal@dream11.com}
\affiliation{%
  \institution{Dream11}
  \city{Mumbai}
  \state{Maharashtra}
  \country{India}
}

\renewcommand{\shortauthors}{Agarwal et al.}

\begin{abstract}


Contextual Bandit (CB) algorithms are widely adopted for personalized recommendations but often struggle in dynamic environments typical of fantasy sports, where rapid changes in user behaviors and dramatic shifts in reward distributions due to external influences necessitate frequent retraining. To address these challenges, we propose a Hierarchical Contextual Uplift Bandit framework. Our framework dynamically adjusts contextual granularity from broad, system-wide insights to detailed, user-specific contexts, using contextual similarity to facilitate effective policy transfer and mitigate cold-start issues. Additionally, we integrate uplift modeling principles into our approach. Results from large-scale A/B testing on the Dream11 fantasy sports platform show our method significantly enhances recommendation quality, achieving 0.4\% revenue improvement, while also improving user satisfaction metrics, compared to the current production system. Subsequently we deployed this system to production as default catalog personalization system in May 2025 and observed a 0.5\% revenue improvement.





\end{abstract}





\maketitle
\section{Introduction}


Daily fantasy sports (DFS) ``contests'' are short-term competitions where participants construct a virtual team comprised of athletes competing in a real world sports match. These teams earn fantasy points based on the statistical performance of the selected athletes in that actual match, and prizes are awarded to the teams that accumulate the most points. For any given match, Dream11--a leading fantasy sports platform with hundreds of millions of users--offers a wide ranging catalog of different contests for users to participate in, from massive low entry amount contests with millions of teams to high stakes head-to-head competitions. 

Our setting is highly dynamic, characterized not only by significant heterogeneity in user preferences but also rapid changes within individual user preferences, say as a result of receiving promotional discounts or winning large prizes. Furthermore, broader ``system-level'' conditions frequently shift, like the start or conclusion of major tournaments, weather delays, or athlete injuries. As a result, being able to adapt and personalize our catalog for a specific context is critical to both sustaining user engagement and platform revenue.

One natural approach would be to leverage a contextual bandit (CB) framework, which balances exploration and exploitation based on real-time user context ~\cite{712192}. Popular algorithms like LinUCB and Thompson Sampling efficiently balance reward estimation and uncertainty, making them suitable for real-time personalization tasks in production settings ~\cite{712192}. Li et al.\cite{li2010contextual} pioneered large-scale applications with LinUCB for personalized news recommendations. Subsequently, CB algorithms have been extensively deployed across various domains, including e-commerce \cite{10.1145/3632410.3632448}), video streaming (e.g., personalized thumbnails on Netflix \cite{10.1145/3240323.3241729}), and large-scale advertising\cite{10.1145/2911451.2911528}. 

Yet, despite their proven effectiveness in large-scale production environments, conventional CB approaches often struggle in highly dynamic settings such as ours, either failing to adapt quickly enough or necessitating frequent re-training, producing far more regret than the stated bounds \cite{luo2018efficient} . Instead, we developed a Hierarchical Contextual Uplift Bandit (HCUB) framework that relies on a hierarchical representation of contextual information \cite{10.5555/3042573.3042700, doi:10.1137/1.9781611975321.74}. This allows our framework to be adaptive in granularity of the personalization, incorporating fine-grained user-specific information when it is meaningful or available, but leveraging broader system-level context otherwise. Moreover, our framework directly integrates uplift modeling principles into the objective function \cite{gutierrez2017causal}, thereby optimizing for incremental gains in user engagement and platform revenue. 


We deployed our framework to directly personalize the catalog. It’s important to note that this is fundamentally different from a recommendation system--rather than just ranking contests already available, we actively change the set of contests that are available for users to join. Using an online A/B test, we piloted our framework on a limited subset of our catalog with a representative sample of 6 million Dream11 users and observed a 0.42\% revenue improvement compared to the BAU system. After this initial result, we deployed the system to production and achieved a 0.51\% revenue improvement. Using the online A/B test data, we ran some additional offline simulations that showed that our framework achieved a 4\% regret improvement with respect to a base case for the representative sample of 6 million users, and a 5\% regret improvement when the system was deployed to production. Given the complexity of the catalog and the scale of operation at Dream11, this 0.51\% revenue impact is a very significant business improvement.


The rest of this paper is organized as follows: Section ~\ref{sec:related_literature} provides an overview of related literature, covering hierarchical bandit algorithms and uplift modeling approaches. Section ~\ref{sec:methodology} presents our proposed methodology. This section covers the  proposed hierarchical contextual structure, the reward inheritance mechanism along with the specific reward function used in this article. Section ~\ref{sec:evaluation} discusses our experimental design and evaluation results, highlighting the outcomes from both online experiments and offline simulations. Finally, Section ~\ref{sec:conclusions} concludes the paper and outlines potential directions for future work.


\section{Related Literature} \label{sec:related_literature}

\subsection{Hierarchical Bandits}
Hierarchical bandits are a specific class of CB algorithms that impose a hierarchical structure on the action or context space, often yielding improvements to scalability or statistical efficiency\cite{10.5555/3042573.3042700}. These gains can be quite substantial: in a bandit setting with millons of arms, Sen et al. \cite{sen2021top} demonstrated that an arm hierarchy can exponentially reduce the number of relevant arms for a given context, making a seemingly intractable problem tractable. Similarly, Hong et al. \cite{hong2022deep} introduced Hierarchical Thompson Sampling (HierTS), which reduces the complexity of a large action space by capturing correlations between action rewards. A key insight from this research is that observing a reward for a specific arm (a leaf node) provides information that can be propagated up the hierarchy, thereby updating beliefs about related arms and parent categories. This notion of sharing rewards across the hierarchy is a powerful mechanism for improving learning efficiency. 

Consequently, hierarchical structures potentially provide a useful way to address the cold-start problem that is common in many real-world settings such as ours. In particular, our framework is largely inspired by Yue et al. \cite{10.5555/3042573.3042700}. Their key innovation is a ``coarse-to-fine'' hierarchical approach, where the model can fall back to using the reward signal of a data-rich parent node reward when the child node reward signal isn't statically meaningful. Our framework leverages a similar reward inheritance mechanism but adapts and extends the hierarchy beyond the user-cohort level to broader system-level context as well. This not only addresses the cold-start problem in our case, but also acts as a regularizer against context fluctuations at the user-cohort level.

\subsection{Causal Bandits}
Though the bandits and causal inference literatures have historically developed largely independently, there is a substantial amount of overlap in terms of the problem spaces and mathematical challenges. Recently, there has been an increasing amount of work that is bridging the gap between the two disciplines, leading to the development of causal bandit algorithms which incorporate insights from causal inference. Some of the developments in this space formulate bandit arms as a node in causal graph , which can help address confounding and lead to lower regret bounds \cite{lattimore2016causal} or improved efficiency in best-arm identification tasks \cite{sen2017identifying}.

Another line of research instead replaces the classical CB regret formulation with one based upon uplift modeling. Rather than measure regret in terms of the loss of expected reward, uplift bandits instead measure regret as the difference in uplift or incrementality between treatment arms \cite{hsieh2022uplifting, pmlr-v67-gutierrez17a}. The distinction here is subtle. For instance, a classical CB might consistently recommend a treatment simply because patients who receive it tend to have better outcomes overall. In contrast, an uplift bandit would target only those patients whose outcomes actually improve as a direct result of receiving the treatment. This property seems especially important in our setting, as there is significant heterogeneity in user behavior that could lead a classical regret formulation down the wrong path (e.g. it starts delivering actions that are associated with high rewards due to power users to the entire population). Moreover, this type of regret formulation has shown significant promise in large scale production settings \cite{geng2021comparison, zhao2022mitigating, kanase2022application}. 

\section{Methodology} \label{sec:methodology}

\subsection{Catalog Representation and Action Space}
\begin{figure}[htbp]
  \centering
  \includegraphics[width=.9\linewidth]{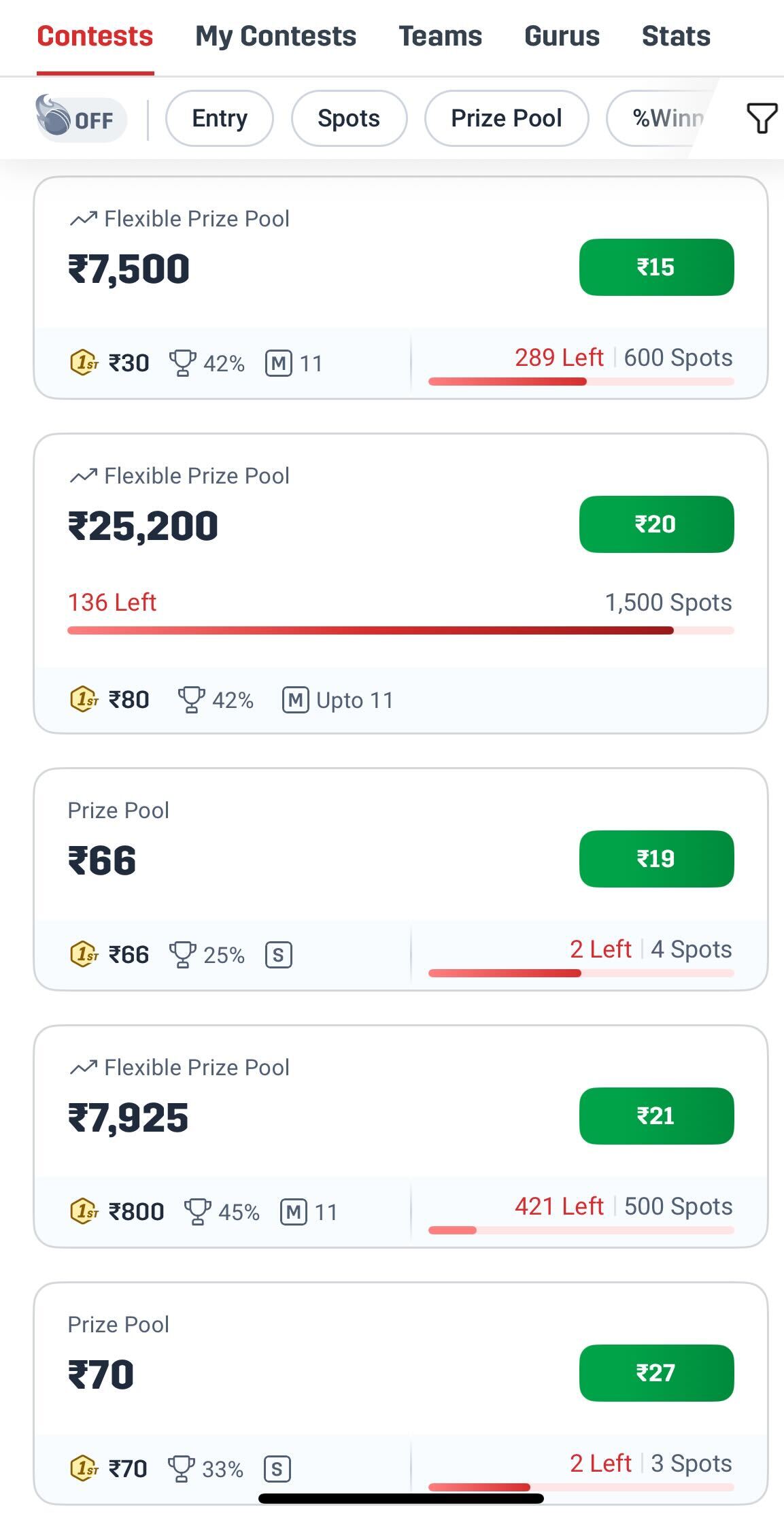}
  \caption{The Contest Selection page on the Dream11 mobile app. It shows the current state of available contests for a particular match. More contests are available by scrolling up or down.}
  \label{fig:app}
\end{figure}


Dream11's current catalog is the result of a combination of many years of product intuition, trial-and-error, and iterative development. Over the years, we have done multiple contest/template-level experiments (not catalog-level). In most of these experiments, we do an online A/B experiment, where we add a set of new templates to our TG and check that the adoption of these new templates is above a certain threshold. Here, the overall business metric (revenue) of the TG is the primary north star metric. It currently consists of $5-10$ contest types and $120-200$ contests in the catalog, depending on the type of the match. See \ref{fig:app} for a snapshot of the contest selection page of a particular match. These contests are composed of 4 main attributes:

\begin{enumerate}
    \item Entry Amount: the amount of money a user needs to pay to participate in the contest. This can range from as little as 1 INR to as high as 25000 INR
    \item Contest Size: the number of teams that can entered into a contest (some contests allow a single user to enter multiple teams). This can range from 2 to 10M+.
    \item Prize Pool: The total amount of money the contest will distribute in prizes. The prize pool of a contest scales with entry amount and contest size.
    \item Prize Distribution: How the prize pool is distributed. Some examples are: winner-take-all (the team with the highest score wins), the top 40\%  double their entry, stepped exponential decay.
\end{enumerate}

Our HCUB framework interfaces with our contest catalog by taking an existing contest and modifying its attributes. Due to the potential risks, the initial deployment of our system operated with major restrictions: it could only modify the contest size attribute, and only for a subset of our catalog. Each of these contests in the catalog subset would now have three variants:
\begin{itemize}
    \item High (\(H\)): a variant of the baseline contest that increases the number of entries
    \item Medium (\(M\)): the baseline contest
    \item Low (\(L\)): a variant of the baseline contest that decreases the number of entries
\end{itemize}

To facilitate robust, tractable, and interpretable exploration, we further grouped the contests in our catalog subset into 4 buckets, such that HML assignment applied to all contests within a bucket. This ultimately leads to an action space of 81 where each arm can be represented by a vector, e.g. \([H, M, M, L]\) or \([M, L, L, L\). We also define the baseline arm to be \([M, M, M, M]\).


\subsection{Reward Function}
We define our reward function as a combination of the uplift of 3 of our key business metrics:
\begin{itemize}
\item Short-term engagement ($Y_1$), which captures direct user activity within a short-window following our framework serving a catalog to the user
\item Long-term retention proxy ($Y_2$), is a metric that we can observe in the short-term but we have found to have a strong established relationship with long-term retention
\item Revenue ($Y_3$), which is simply the total of a users' entry amount on the platform
\end{itemize}
We formally define the uplift of a metric as follows:

\begin{equation}
    \Delta_{Y_k}(a) = Y_k(A = a) - Y_k(A = [M, M, M, M])
    \label{eq:uplift}
\end{equation}

Or the difference in value between of a particular metric given the action vector that was actually delivered vs the baseline action vector. Our final reward $R$ is simply a weighted sum of these uplifts:

\begin{equation}
    R(a) = \lambda_1 \cdot \Delta Y_1(a) + \lambda_2 \cdot \Delta Y_2(a) + \lambda_3 \cdot \Delta Y_3(a)
    \label{eq:reward}
\end{equation}

where weights $\lambda_1, \lambda_2, \lambda_3$ are values determined by strategic business requirements.

\subsection{Hierarchical Context and Reward Inheritance}\label{subsec:reward-inheritance}
\begin{figure*}[htbp]
  \centering
  \includegraphics[width=0.7\linewidth]{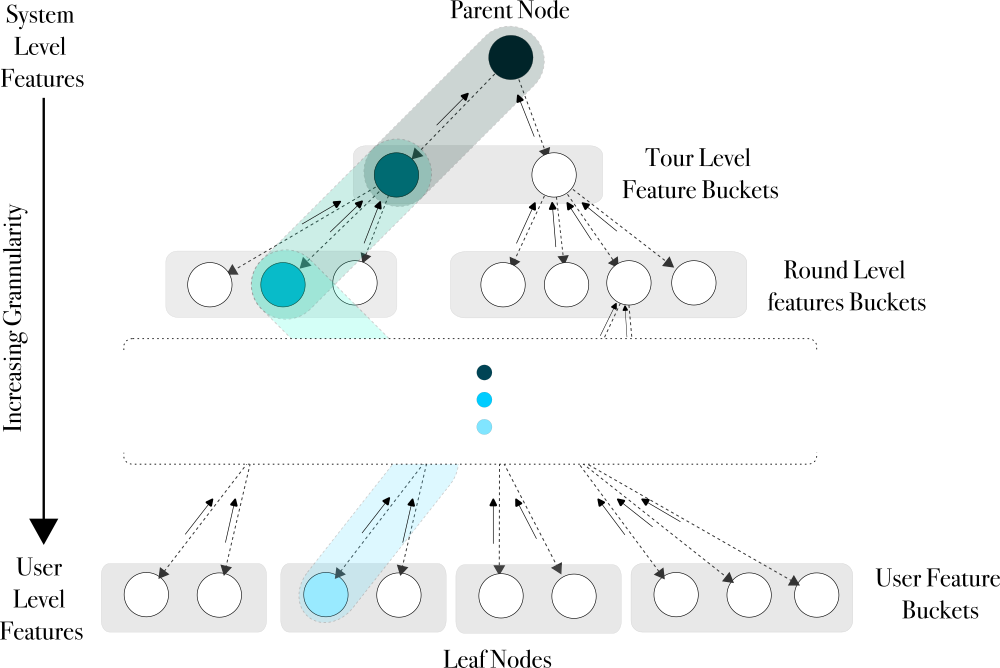}
  \caption{Hierarchical Contextual Tree representation illustrating the relationship between system-level and user-level features in contextual bandits. The root node represents system-level features, with branches corresponding to distinct cohorts. Feature granularity increases along the depth of the tree, culminating in user-level features at the leaf nodes. The shaded regions highlight an example context path from the root to leaf nodes. The uplift for each context-action pair are computed relative to a base action. Uplift values calculated at leaf nodes are aggregated upwards (represented by the solid arrows), while rewards propagate downward from parent to child nodes (represented by the dotted arrows).}
  \label{fig:context-tree}
\end{figure*}

\begin{algorithm}[htbp] \label{algo}
\SetAlgoLined
\KwIn{Hierarchical context tree $\mathcal{T}$, baseline action $(a_{BAU})$, action set $\mathcal{A}$, contexts $X$, rounds $T$}
\KwOut{Recommended action $a_t$ at each round $t$ for a given user(context X)}

\textbf{Algorithm}: \\
Initialize hierarchical context tree $\mathcal{T}$\;
Define baseline action $a_{BAU}$ and alternative actions $\mathcal{A}$\;

\ForEach{recommendation round $t=1,\dots,T$}{
    Observe user context $x_t$ and identify leaf node $N$\;
    
    \tcp{Bottom-up aggregation of uplift estimates}
    \ForEach{leaf node $n\in \mathcal{T}$}{
        \ForEach{action $a\in\mathcal{A}\setminus\{a_{BAU}\}$}{
            Compute $uplift(n,a)) $ via bootstrap sampling\;
            Compute confidence interval $CI(n,a)$\;
        }
    }

    \ForEach{internal node $i$ from leaves to root}{
        \ForEach{action $a\in\mathcal{A}\setminus\{a_{BAU}\}$}{
            Aggregate uplifts from child nodes:
            $uplift(i,a)\leftarrow\sum_{child\in i}{w_{child}\cdot uplift(child,a)}$\;
            $CI(i,a)\leftarrow$ aggregate confidence intervals from children\;
        }
    }

    \tcp{Top-down reward inheritance}
    \tcp{Remove for offline simulations}
    \ForEach{node $n$ from root to leaves}{
        \ForEach{action $a\in\mathcal{A}\setminus\{a_{BAU}\}$}{
            \eIf{$CI(n,a)$ indicates statistically significant uplift}{
                $uplift(n,a)\leftarrow uplift(n,a)$\;
            }{
                $uplift(n,a)\leftarrow uplift(parent(n),a)$\;
            }
        }
    }

    \tcp{Action selection at leaf node using Bayesian UCB}
    \ForEach{action $a\in \mathcal{A}$}{
        Compute Bayesian UCB score at node $L$: \\
        $UCB\_score(a)\leftarrow uplift(L,a)+exploration\_term(L,a)$\;
    }
    
    Select action: $a_t\leftarrow\arg\max_{a\in\mathcal{A}}{UCB\_score(a)}$\;
    Apply action $a_t$ and observe reward $r_t$\;
    Update reward estimates based on $(x_t,a_t,r_t)$\;
}
\caption{Hierarchical Contextual Uplift Bandit (HCUB)}
\label{alg:hcub}
\end{algorithm}

The contextual information our framework uses encompasses both fine-grained user data as well as characteristics of the upcoming matches. We impose a hierarchical tree structure on this contextual information (illustrated in Figure~\ref{fig:context-tree}). Near the top, we have internal nodes defined features related to a ``tour'' which is a collection of matches such as the Indian Premier League (Cricket), NBA (Basketball), or the English Premier League (Football/Soccer); or by features related to a particular match or ``round'', e.g. which teams are playing, what time is the match, etc. At the bottom of the hierarchy, we have leaf nodes that are defined user-level attributes such as predicted churn score, recent transactions, historical winning behavior, participation across contest types, and preferences for particular contest attributes. 


Our HCUB framework calculates the uplift on our joint reward metric for each leaf node-arm. Reward inheritance subsequently occurs in a top-down manner (represented by dotted arrows in Figure~\ref{fig:context-tree}). Starting from the root, optimal reward values are propagated downward through the hierarchy. If a child node demonstrates statistically significant uplift relative to the baseline arm, it retains this calculated reward. Otherwise, it inherits the reward from its immediate ancestor recursively until a statistically significant reward is identified or the root is reached. This top-down inheritance mitigates the cold-start problem, as ancestors have richer data due to larger sample sizes, ensuring more reliable estimates and synchronized exploration across hierarchy levels.

To accurately quantify uncertainty and dependencies between related user cohorts, hierarchical bootstrapping is utilized. Specifically, for each node-action pair, multiple subsamples are drawn separately for the treatment action and the baseline action. From these samples, uplift scores relative to baseline are computed, yielding median treatment effects and confidence intervals (5th and 95th percentiles). These estimates are aggregated upward in a bottom-up fashion through internal nodes (represented by solid arrows in Figure~\ref{fig:context-tree}), proportionally accounting for the size of each node's child cohorts. After this step, each node-action pair across the tree possesses calculated uplift scores and confidence intervals.


Reward inheritance subsequently occurs in a top-down manner (represented by dotted arrows in Figure~\ref{fig:context-tree}). Starting from the root, uplifts are propagated downward through the hierarchy. If a child node demonstrates statistically significant uplift for an arm (at a $95\%$ confidence level), it retains this calculated reward. Otherwise, it inherits the reward from its immediate ancestor recursively until a statistically significant reward is identified or the root is reached. This top-down inheritance mitigates the cold-start problem, as ancestors have richer data due to larger sample sizes, ensuring more reliable estimates and synchronized exploration across hierarchy levels.

Finally, arm selection is performed bottom-up via Bayesian Upper Confidence Bound (UCB). Each leaf node selects the arm (catalog configuration) with the highest posterior upper confidence bound, informed by its own reward estimates and any inherited information. This Bayesian UCB approach efficiently balances exploration and exploitation, particularly beneficial in sparse-data settings, by leveraging statistical strength across related user segments and prioritizing exploration of actions with the highest potential uplift.

\section{Evaluation} \label{sec:evaluation}
We evaluate the effectiveness of our proposed framework, specifically investigating the impact of reward inheritance on personalized recommendation and speed of adaptation in dynamic fantasy sports environments. To this end, we conducted both online experiments and offline simulations. 

Algorithm ~\ref{alg:hcub} outlines the detailed steps of our Hierarchical Contextual Uplift Bandit (HCUB) framework. Initially, uplift scores and corresponding confidence intervals for each leaf-node and action pair are calculated via bootstrapping and aggregated upwards through the hierarchy. Reward inheritance subsequently occurs in a top-down manner (represented by dotted arrows in Figure~\ref{fig:context-tree}). Starting from the root, uplifts are propagated downward through the hierarchy. If a child node demonstrates statistically significant uplift for an arm (at a $95\%$ confidence level), it retains this calculated reward. Otherwise, it inherits the reward from its immediate ancestor recursively until a statistically significant reward is identified or the root is reached. This top-down inheritance mitigates the cold-start problem, as ancestors have richer data due to larger sample sizes, ensuring more reliable estimates and synchronized exploration across hierarchy levels.

\subsection{Online Experiments}

The online evaluation involved large-scale A/B testing conducted in two stages. The first experiment engaged a representative sample of 6 million Dream11 users from February-March 2025. Subsequently, this system was scaled up for the whole Dream11 population starting May 2025. In both stages, users's contest catalog was updated at fixed time intervals, based on evolving user context, and performance was compared against the existing production catalog. These online experiments utilized the reward inheritance framework described in Section ~\ref{subsec:reward-inheritance}, with the reward function defined as shown in Equation ~\ref{eq:reward}. For online evaluation, we prioritized revenue ($\lambda_3 = 1.0$) while still balancing short-term engagement ($\lambda_1 = 0.5$) and long-term retention ($\lambda_2 = 0.5$).

\subsection{Offline Simulations}
Utilizing data collected from the online experiments, we further conducted offline simulations to isolate and explicitly evaluate the impact of reward inheritance. In our offline experiments, we employed the same Hierarchical Contextual Uplift Bandit (HCUB) algorithm as described in Algorithm~\ref{alg:hcub}, with one key modification: we disabled reward inheritance from ancestors by removing the \textbf{foreach} loop below the \textbf{Top-down reward inheritance} comment. This adjustment allowed us to explicitly evaluate the impact of reward inheritance by directly comparing the original algorithm against a variant without hierarchical inheritance. Using this simulation, we compared the results of the HCUB framework with reward inheritance (the proposed online experiment setup) with the same HCUB setup without the reward inheritance, enabling a direct assessment of the benefits introduced by our approach.

\subsection{Results}



In offline simulations, our framework achieved a 4\% regret improvement in the first simulation scenario and a 5\% regret improvement in the second scenario when utilizing reward inheritance, compared to not using inheritance, while keeping all other factors constant.

Table~\ref{tab:experiment-results} summarizes the outcomes of the online experiments. We observed significant positive improvements in key financial metrics, specifically significant  revenue uplift was observed during both phases of the experiments. Although our engagement metric, DAU (Daily Active Users) showed slight uplifts, these were statistically insignificant.  These findings indicate a consistent and reliable revenue uplift from the experiment, while  maintaining short term and long term user engagement metrics at-least as good as the incumbent production catalog system.

\begin{table}[htbp]
\centering
\caption{Online Experiment Results}
\label{tab:experiment-results}
\renewcommand{\arraystretch}{1.5}
\begin{tabular}{| l | p{3.5cm} p{3cm} |}
\hline\hline
\textbf{Metric} & \textbf{Feb -- Mar 2025\newline (6M users)} & \textbf{May -- Present\newline (Entire D11\newline user base)} \\[1ex]
\hline
Revenue & \textbf{+0.42\%} ($p=0.041$)& \textbf{+0.51\%} ($p=0.034$) \\[0.8ex]
DAU & +0.05\% (insignificant) & +0.1\% (insignificant) \\[0.8ex]
\hline\hline
\end{tabular}
\end{table}

\section{Conclusions and Future Work} \label{sec:conclusions}
In this paper, we proposed a Hierarchical Contextual Uplift Bandit framework that leverages an uplift-based reward, which is inherited hierarchically across a context tree. While we applied this approach on the problem of catalog personalization in fantasy sports, it should be able to generalize well to other personalization tasks in other highly dynamic environments such as video streaming or e-commerce.

We piloted this framework on our production system, where it actively personalized a subset of our catalog for over 6M users. Despite the major limitations we imposed our approach, our online A/B test still showed a significant improvement in our joint reward metric. We are currently in the process of scaling up our framework to cover additional contest attributes, as well as cover a larger subset of our catalog.

Beyond just expanding the scope of the deployment, we are also looking into further improvements to the core framework. One promising area we are exploring is in the space of adaptive bandits, which employ methods such discounting, sliding windows, and change-point detection to address non-stationarities.

\bibliographystyle{ACM-Reference-Format}
\bibliography{sample-base}

\end{document}